\documentclass{article}

\usepackage{PRIMEarxiv}

\usepackage{graphicx}
\usepackage{booktabs}
\usepackage{multirow}
\usepackage{multicol}
\usepackage{makecell}
\usepackage{epsfig}
\usepackage[utf8]{inputenc} 
\usepackage[T1]{fontenc}    
\usepackage{amsmath}
\usepackage{adjustbox}
\usepackage{hyperref}       
\usepackage[capitalize]{cleveref}
\usepackage{url}            
\usepackage{booktabs}       
\usepackage{amsfonts}       
\usepackage{nicefrac}       
\usepackage{microtype}      
\usepackage{lipsum}
\usepackage{fancyhdr}       
\usepackage{graphicx}       
\graphicspath{{media/}}     
\usepackage{enumitem}

\usepackage{algorithm}
\usepackage{algorithmicx}
\usepackage{algpseudocode}
\usepackage{bm}

\usepackage{rotating}
\usepackage{gensymb}
\usepackage{wrapfig}
\usepackage{xcolor}
\usepackage{array}
\usepackage{colortbl}
\usepackage{pifont}
\definecolor{mygray}{gray}{.92}

\newcommand{\cmark}{\ding{51}}
\newcommand{\xmark}{\ding{55}}

\definecolor{yzycolor}{rgb}{0.96,0.57,0.58}

\definecolor{sytcolor}{rgb}{0.5,0.67,0.89}

\definecolor{best}{RGB}{234, 239, 239}
\definecolor{second}{rgb}{0.98, 0.78, 0.57}
\definecolor{third}{rgb}{1.0, 1.0, 0.56}

\usepackage{xspace}
\newcommand{\model}{FlowGaussian-VR\xspace}    

\title{Enhanced Velocity Field Modeling for Gaussian Video Reconstruction}

\author{
  Zhenyang Li\textsuperscript{1}\thanks{These authors contributed equally.},
  Xiaoyang Bai\textsuperscript{1}\footnotemark[1],
  Tongchen Zhang\textsuperscript{2},
  Pengfei Shen\textsuperscript{1},
  Weiwei Xu\textsuperscript{2},
  Yifan Peng\textsuperscript{1}\thanks{e-mail: evanpeng\texttt{@}hku.hk} \\[1ex]
  \scriptsize
  \textsuperscript{1}The University of Hong Kong\quad
  \textsuperscript{2}State Key Laboratory of Computer-aided Design \& Computer Graphics, Zhejiang University
}

\begin{document}
\maketitle

\begin{abstract}
    High-fidelity 3D video reconstruction is essential for enabling real-time rendering of dynamic scenes with realistic motion in virtual and augmented reality (VR/AR).
    The deformation field paradigm of 3D Gaussian splatting has achieved near-photorealistic results in video reconstruction due to the great representation capability of deep deformation networks. 
    However, in videos with complex motion and significant scale variations, deformation networks often overfit to irregular Gaussian trajectories, leading to suboptimal visual quality. 
    Moreover, the gradient-based densification strategy designed for static scene reconstruction proves inadequate to address the absence of dynamic content.
    In light of these challenges, we propose a flow-empowered velocity field modeling scheme tailored for Gaussian video reconstruction, dubbed \emph{\model}. It consists of two core components: a \emph{velocity field rendering (VFR)} pipeline which enables optical flow-based optimization, and a \emph{flow-assisted adaptive densification (FAD)} strategy that adjusts the number and size of Gaussians in dynamic regions. 
    We also explore a \emph{temporal velocity refinement (TVR)} post-processing algorithm to further estimate and correct noise in Gaussian trajectories via extended Kalman filtering.
    We validate our model's effectiveness on multi-view dynamic reconstruction and novel view synthesis with multiple real-world datasets containing challenging motion scenarios, demonstrating not only notable visual improvements (over 2.5~dB gain in PSNR) and less blurry artifacts in dynamic textures, but also regularized and trackable per-Gaussian trajectories.
\end{abstract}


\begin{figure}
    \centering
    \includegraphics[width=\linewidth]{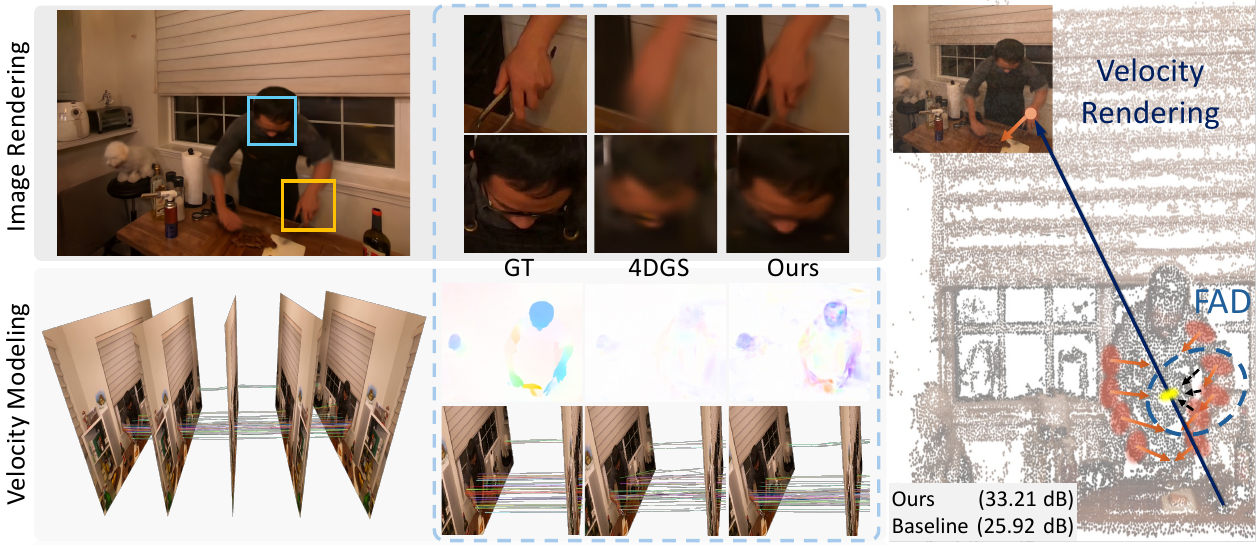}
    \caption{\textbf{Left}: Deformation-based 4DGS encounters difficulties in reconstructing scenes and rendering novel views under challenging conditions, such as significant motion and other complex dynamics. Our \model exhibits commendable performance on given scenes (e.g., ``cut-roasted-beef''). \textbf{Middle}: We compare the ground-truth optical flow, the deformation network of baseline (4DGS), and the velocity field rendered by our method. \textbf{Right}: We render the velocity field for scene Gaussians, constrain it with flow-based losses, and employ the FAD strategy to add Gaussians for dynamic objects in the canonical space.}
    \label{fig:teaser}
\end{figure}

\begin{figure*}[t]
  \centering
   \includegraphics[width=0.995\linewidth]{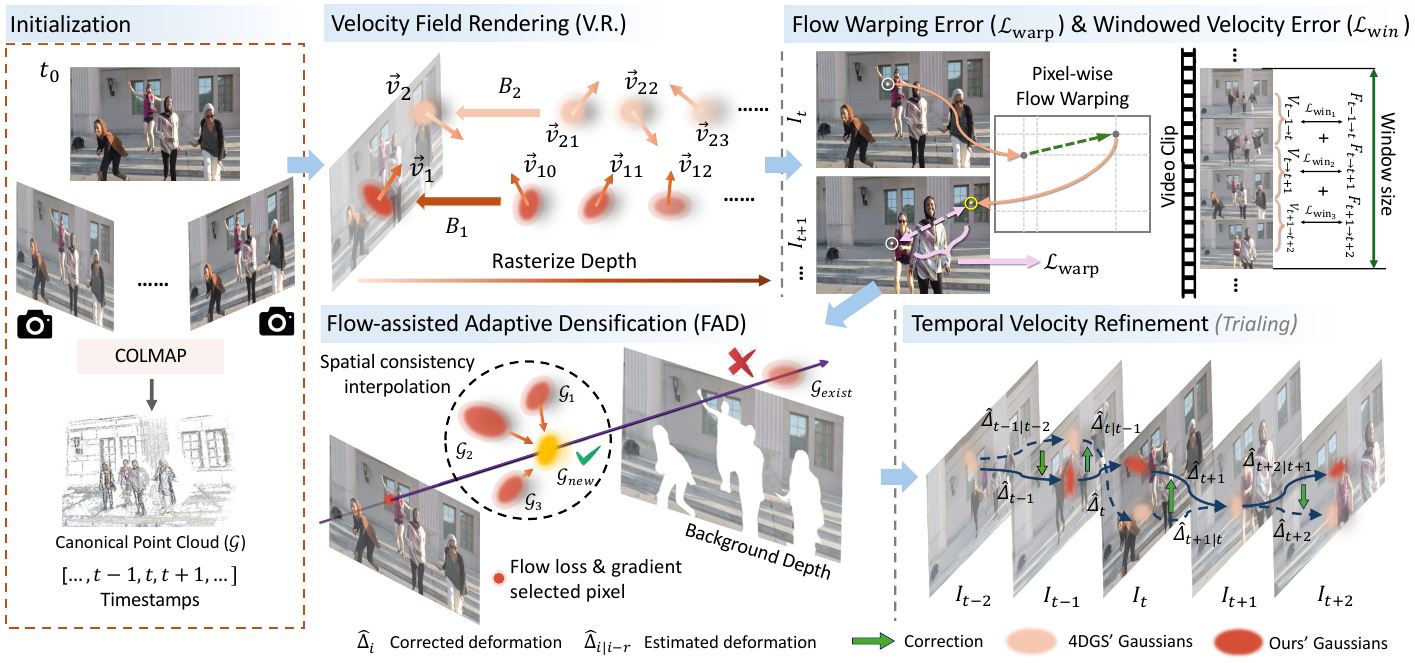}
   \caption{\textbf{Overview of our \model pipeline}. The velocity attribute is introduced into each Gaussian within the dynamic 3D GS pipeline to ensure that the motion of Gaussians closely mimics real-world physical motion. To achieve this, we impose several constraints on the rendered velocity: $\mathcal{L}_{\mathrm{warp}}$ ensures temporal color consistency by aligning the predicted velocity with sequential frame data; $\mathcal{L}_{\mathrm{win}}$ maximizes the supervision effect of ground truth optical flow on velocity predictions; $\mathcal{L}_{\mathrm{dyn}}$ enhances rendering quality and velocity accuracy for dynamic regions. Finally, to globally refine the predicted velocity across the temporal domain, we explore a TVR post-processing strategy, which integrates the traditional Extended Kalman Filter (EKF).
   }
   \label{fig:pipeline}
\end{figure*}

\section{Introduction} 
\label{sec:introduction}
3D scene reconstruction is a crucial step for crafting immersive and realistic experiences in a wide range of augmented and virtual reality (AR/VR) applications~\cite{ismar-1,ismar-4}, including cinema-level virtual modeling~\cite{jiang2024vrgs,chen2024dreamcinema}, autonomous driving~\cite{zhou2024drivinggaussian}, robotics~\cite{pan2024robust,hu2024hoimotion}, and medical surgery~\cite{dastan2024co}, as it accurately represents the spatial layouts of 3D environments and readily enables users to interact with and even manipulate scene contents.
In this domain, 3D Gaussian Splatting (3DGS)~\cite{kerbl20233d} has recently demonstrated remarkable progress in both reconstruction quality and rendering efficiency, outperforming established neural radiance field (NeRF) variants~\cite{original.nerf}. 
Noteworthy progress has also been observed in the more challenging dynamic scene reconstruction task~\cite{luiten2023dynamic3dgs,li2024spacetime,wu20244d}, where 
Gaussian Splatting (GS) is also being increasingly adopted to modeling dynamic scenes. Existing dynamic GS frameworks mainly follow two schemes: \emph{deformation-based} methods learn a deformation field~\cite{Wu_2024_CVPR} to represent 3D motions, while \emph{trajectory-based} methods explicitly model the geometry transformation between consecutive frames~\cite{gao2024gaussianflow,wang2024shape}. 
The former scheme, leveraging the exceptional representation capability of neural networks, has generally demonstrated better reconstruction quality across diverse datasets.

We attribute the superior rendering quality of deformation-based methods to two key factors: minimal motion between frames which facilitates model convergence, and large Gaussian population which yields sufficient optimizable parameters during training.
However, as shown in~\cref{fig:teaser}, Gaussian trajectories predicted by the deformation network often lack local consistency and struggle to align with the true dynamics of moving objects, leading to suboptimal video reconstruction quality and restricting the application of these methods to high-quality video datasets with fewer camera fluctuations, small-scale motion, and clear content.
Stable and physically accurate Gaussian trajectories ensure both rendering and semantic consistency in dynamic regions during novel view synthesis, resulting in photorealistic outcomes.
Therefore, their prospects in fields such as autonomous driving and robotics, where critical applications in semantic segmentation and object tracking demand accurate reconstruction of object trajectories~\cite{huang2022survey}, appear limited.

As optical flow~\cite{zhu2023nicerslam} has been proven effective in enhancing the camera pose estimation consistency in monocular video reconstruction, we adopt a similar approach to supervising the motion of 3D Gaussians. Nonetheless, we observe that such a methodology cannot be trivially applied to Gaussian representations for two primary reasons.
Firstly, the centers of 3D Gaussians do not align with object surfaces. This misalignment may cause conflicts between velocity field-based learning and photometric supervision, leading to misguided optimization.
Secondly, the default densification strategy of dynamic GS, inherited from 3DGS, relies on the gradient of photometric losses~\cite{luiten2023dynamic3dgs}. We find experimentally that this strategy struggles to capture regions with substantial motion and does not effectively align with the dynamic information present in the scene.
Therefore, how to overcome these problems and incorporate motion cues into GS frameworks becomes a critical challenge.

To this purpose, we propose \textit{\model}, a velocity field modeling pipeline that incorporates optical flow as the ground truth for 3D Gaussian dynamics to enhance their temporal controllability and interpretability. We render 2D velocity fields with differentiable rasterization for each Gaussian and supervise them using outputs from models such as RAFT~\cite{teed2020raft}. Consequently, we introduce three losses based on the rendered velocity fields to facilitate the alignment of Gaussian attributes with real-world dynamics captured by estimated optical flows. The \textit{windowed velocity error ($\mathcal{L}_{\mathrm{win}}$)} employs a temporal sliding window to optimize rendered velocity fields across multiple frames. The \textit{flow warping error ($\mathcal{L}_{\mathrm{warp}}$)} computes the discrepancy between rendered images and warped ground truths from the next frame. Additionally, we leverage SAM-v2~\cite{ravi2024sam2} to segment dynamic regions for each video and compute the \textit{dynamic rendering loss ($\mathcal{L}_{\mathrm{dyn}}$)} to further refine the model's representation of dynamic scene components.

While velocity field-based optimization effectively controls the trajectories of Gaussian centers, the combination of flow-based losses and the original photometric loss may still fall behind object motion when the scene undergoes significant changes between frames.
To mitigate this issue, we introduce a \textit{flow-assisted adaptive densification (FAD)} strategy that adds Gaussians by identifying challenging dynamic regions on the 2D frame instead of cloning and splitting existing ones.
Compared to the conventional densification algorithm, this design helps to complete scene components that are missing from initialization based on a single video frame. 
Finally, we explore the possibility of leveraging the Extended Kalman Filter (EKF)~\cite{fujii2013ekf} as a post-processing \emph{temporal velocity refinement (TVR)} technique for Gaussian motion estimation.

We conduct a comprehensive evaluation of our model on challenging datasets: Nvidia-long~\cite{yoon2020dynamicnvs}, and Neu3D~\cite{li2022neu3d}. 
Experimental results indicate that our \model pipeline, when integrated with the 4DGS~\cite{Wu_2024_CVPR} baseline, improves the overall PSNR of novel view rendering by approximately 2.5~dB across challenging scenes, with a gain of over 2.0~dB on dynamic regions. Additionally, we compare our method with alternative approaches following the same scheme, namely 4D-GS~\cite{yang2023gs4d} and SC-GS~\cite{huang2024scgs}, yielding superior performance.
We further demonstrate that the TVR strategy effectively smoothens Gaussian trajectories and aligns them more closely with ground truth optical flows, substantiating the potential of a multi-stage learning framework for dynamic GS.

In summary, our contributions are as follows:
\begin{itemize}[leftmargin=12pt,topsep=2pt,itemsep=2pt]
    \item We propose a velocity field modeling pipeline, \emph{\model}, which facilitates optical flow-based optimization via velocity field rendering in dynamic Gaussian splatting models to enhance the temporal behavior of Gaussian instances.
    \item We introduce the FAD strategy to address challenges of optimizing Gaussians in dynamic scenes with abrupt changes. By integrating flow-based supervision into its pipeline, FAD enhances the  robustness of dynamic GS video processing.
    \item We select challenging datasets comprising real-world captured videos to experimentally demonstrate the superior performance of \model on complex dynamic scenes.
\end{itemize}
\section{Related Work} \label{sec:related work}
\subsection{Video Reconstruction}
Neural video reconstruction methods~\cite{wang2022neural,bozic2020deepdeform} have significantly advanced the modeling and rendering of dynamic scenes~\cite{siarohin2019first, dou2016fusion4d}, with progress in monocular videos~\cite{kopf2021robust, zhang2021consistent}, time-synchronized videos~\cite{Kim2024videosync}, sparse camera views~\cite{wang2023sparsenerf}, and stereo cameras~\cite{Tosi_2023_nerfstereo}.
In dynamic settings, NeRF-based approaches have made notable strides, particularly for video reconstruction. Some models represent dynamic scenes as time-conditional radiance fields~\cite{du2021neural,tretschk2021non,xian2021space}, or use deformation networks separate from the canonical radiance field~\cite{pumarola2021d}. Others, such as DyNeRF~\cite{li2022neu3d}, Nerfies~\cite{park2021nerfies}, and HyperNeRF~\cite{park2021hypernerf}, encode temporal information into latent representations. 
For trajectory learning, approaches like DynamicNeRF~\cite{gao2021dynamic} and NSFF~\cite{li2021neural} regress optical flow from spacetime coordinates, while DynIBaR~\cite{li2023dynibar} models trajectory fields using DCT bases. However, NeRF representations are computationally intensive, posing challenges for reconstructing complex dynamic scenes in videos.

\subsection{Dynamic 3D Gaussian Splatting}
The emergence of 3DGS~\cite{kerbl20233d,ren2024octree,yu2024mip} has significantly improved training speeds and reduced memory usage in video reconstruction models. Dynamic 3DGS variants primarily fall into two categories: those with time-dependent Gaussian parameters~\cite{luiten2023dynamic3dgs,li2024spacetime} and those employing deformation fields~\cite{yang2024deformable,wu20244d}. The latter offers a lightweight network, an initial point cloud, and fast training. These advantages make them well-suited for 3D video streaming, compression, and web-based demonstrations, while directly applying the previous type of dynamic 3DGS in these applications may struggle to handle the substantial storage concern.
However, many 3DGS methods rely on datasets with minimal motion blur, slow object motion, and without flickers. When interference factors are present, dynamic 3DGS~\cite{luiten2023dynamic3dgs} performs multiview 3D optimization on the point clouds of each frame, but the high number of iterations significantly increases training time. A more efficient approach applies deformation to the initial frame instead.  
Yet, optimizing the deformation field solely based on point cloud and temporal features~\cite{cao2023hexplane} lacks explicit constraints~\cite{3d-arap}, reducing its controllability and adherence to physical motion laws. Without well-defined motion dynamics, the model struggles to accurately capture per-instance motion within the point cloud.

\subsection{Flow-assisted Gaussians}
Several studies~\cite{guo2024motion,zhu2024motiongs,xu2024motionrepresenting,wang2024gflow} have explored integrating dynamic flows, such as velocity fields, with dynamic GS. MODGS~\cite{liu2024modgs} leverages off-the-shelf depth~\cite{fu2025geowizard} and flow~\cite{teed2020raft} estimators to compute the 3D flow between frames, initializing the deformation network for point cloud prediction. However, it is constrained by the limitations of depth estimation and struggles to adapt to Gaussian changes during densification and pruning, reducing the effectiveness of flow supervision.
Gaussian-Flow~\cite{gao2024gaussianflow} and Shape of Motion~\cite{wang2024shape} improve temporal motion modeling by introducing motion bases into the deformation network but avoid directly supervising Gaussian parameters or trajectories. These methods also overlook the impact of deformation updates on the rasterizer’s rendering, as Gaussian attributes are indirectly modified through geometric reshaping~\cite{tosi2024nerfs} rather than being directly controlled.
Allowing the rasterizer to render velocity fields provides direct supervision of Gaussian properties, enhancing both deformation controllability and temporal motion representation.
More relevant to our research is MotionGS~\cite{zhu2024motiongs}, which decouples foreground Gaussians from background ones and supervises object motion with pseudo 2D optical flow. However, we observe in practice that these Gaussians still undergo large, erratic displacements during optimization, causing numerous background Gaussians to drift into the foreground and corrupt its rendering. Furthermore, MotionGS only controls the center position of each Gaussian, while overlooking and constraining the optimization space of other attributes. Such a design prevents it from reaching globally optimal solution, especially in fast-moving scenes. Our method overcomes this drawback by directly rasterizing velocity fields from dynamic scenes.
\vspace{-0.2cm}

\begin{figure}[t]
  \centering
   \includegraphics[width=0.45\linewidth]{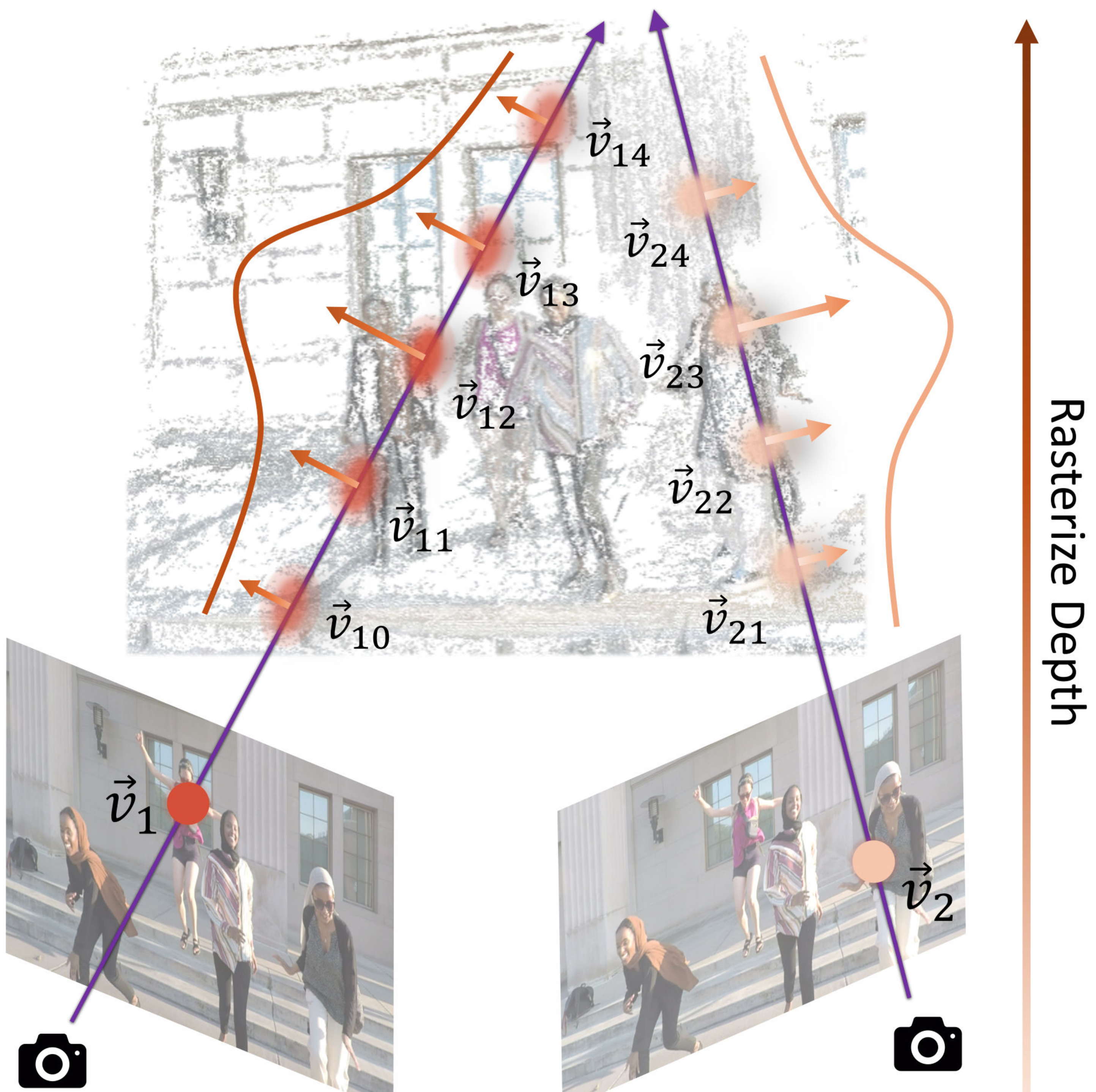}
   \caption{\textbf{Velocity Rendering in 3D Gaussian Rasterization.} Gaussian velocities are  alpha-blended  along each projection ray. Through multi-view optimization, these Gaussian velocities progressively converge toward the actual motion of the object.}
   \label{fig:velocity rendering}
\end{figure}

\section{Gaussian Video with Velocity Field Modeling} 
\label{sec:method}

As evidenced in \cref{fig:teaser}, although deformation-based dynamic GS methods such as 4DGS~\cite{Wu_2024_CVPR} can generate visually appealing rendering results, their per-Gaussian motion estimation lacks constraints and local coherence.
In light of this, we introduce \model, whose pipeline is illustrated in \cref{fig:pipeline}. It comprises a dynamic GS backbone (Sec.~3.1), multiple flow-based constraints empowered by velocity field rendering (Sec.~3.2 \& 3.3), and the flow-assisted adaptive densification strategy (Sec.~3.4).
The proposed pipeline centers around the allocation of 2D velocity attributes (e.g., 2D velocity vector) to each Gaussian point and the subsequent modeling of motion dynamics with this enhanced mapping. We name this mechanism \emph{velocity field modeling} throughout the manuscript.
In addition, we discuss our exploration on the Extended Kalman Filter (EKF)-aided temporal velocity refinement post-processing in Sec.~5 and the supplementary material.

\subsection{Preliminary: 4D Gaussian Splatting (4DGS)} 
\label{subsec: prelim}

In 4DGS~\cite{Wu_2024_CVPR}, a dynamic 3D scene is represented by a set of Gaussians $\mathcal{G}$. Each Gaussian in the scene is defined as $G_i = \left( \mathbf{\mu}_i, \Sigma_i, c_i, \sigma_i \right)$, with $\mathbf{\mu}_i \in \mathbb{R}^3$ being the center coordinates, $\Sigma_i \in \mathbb{R}^{3 \times 3}$ being the covariance matrix controlling the spread and orientation, $c_i \in \mathbb{R}^3$ being the color representation, and $\sigma_i \in [0, 1]$ being the opacity value. For simplicity of presentation, we omit the Gaussian index $i$ hereafter.
To capture temporal changes, 4DGS applies a deformation field $D(\mathbf{x}, t): \mathbb{R}^3\times\mathbb{R}\rightarrow\mathbb{R}^3$ to each Gaussian center $\mathbf{\mu}_0$ in the canonical space, predicting its position at time $t$ as $\mathbf{\mu}_t = \mathbf{\mu}_0 + D(\mathbf{\mu}_0, t)$. The deformation model $D$ is trained end-to-end with the Gaussian parameters using image losses. During rendering, $\mu_t$ is projected onto the 2D plane as $\mu_t'$ and the intensity at pixel $\mathbf{p}$ is computed following the formulation in 3DGS:
\begin{align} \label{eq:4dgs rendering}
    \begin{split}
        I(\mathbf{p}) &= \sum_{i=1}^{N} \alpha_i \, c_i \, \prod_{j=1}^i (1 - \alpha_j), \\ \text{ where } \,
        \alpha_i &= \sigma_i \cdot \exp \left[ -\frac{1}{2} (\mathbf{p} - \mathbf{\mu}_i')^T \Sigma_i^{-1} (\mathbf{p} - \mathbf{\mu}_i') \right].
    \end{split}
\end{align}
Inspired by the alpha blending technique in Eq.~\ref{eq:4dgs rendering}, we propose the velocity field rendering algorithm below to obtain 2D representations of scene motion through differentiable rasterization.



\subsection{Velocity Field Rendering} \label{subsec:velocity fieid rendering}
Given the point cloud $\mathbf{P}_0$ reconstructed by COLMAP~\cite{schonberger2016colmap} and two adjacent timestamps $\{t, t+\delta t\}$,
we obtain the velocity field in a manner similar to the pixel color rendering in \cref{eq:4dgs rendering} by projecting the deformed point clouds, $\mathbf{P}_t$ and $\mathbf{P}_{t+\delta t}$, onto the pixel plane, and calculating their displacement over $\delta t$.

Specifically, we first transform a 3D point cloud $\mathbf{P}$ from world coordinates to camera coordinates by applying the camera's extrinsic parameters, i.e. the rotation matrix $\mathbf{R}$ and translation vector $\mathbf{T}$. 
Then, we project the 3D camera coordinates onto the 2D image plane by utilizing the camera's intrinsic matrix $\mathbf{K}$, which depends on the focal lengths and optical center of the camera. 
Ultimately, the 2D pixel coordinates $\mathbf{p}$ is obtained by normalizing the homogeneous coordinates.
%
In such a way, we enable the rendering of 2D velocity fields from the velocity attributes of Gaussians and the subsequent optimization of Gaussian parameters to capture motion dynamics effectively, completing the velocity field modeling pipeline.

\textbf{Velocity Field Rasterization.} Each Gaussian in the 3D point cloud is assigned a velocity vector that represents its movement between adjacent frames (\cref{fig:velocity rendering}). The projected 2D velocity vector allows us to rasterize the temporal motion information at each pixel, akin to color rasterization in the original 3DGS formulation, and to enable a more explicit control of Gaussian attributes by the deformation network.
Analogous to \cref{eq:4dgs rendering}, the velocity field $V(\mathbf{p})$ rendering process is formulated as:
\begin{align}
    V(\mathbf{p}) &= \sum_{i=1}^{N} \alpha_i \, v_i \, \prod_{j=1}^i (1 - \alpha_j).
\end{align}
By definition, the projected velocity serves as a reasonable estimation of the rendered optical flow, assuming that a pixel is predominantly influenced by Gaussians that contribute to its color, and its radiance is carried away by the motion of those Gaussians. 
Therefore, the velocity field on the pixel is a sum of the velocity of dominant Gaussians, weighed by their densities and visibilities. 
Alternatively, we may render two consecutive frames and apply optical flow estimation on the outcome directly. However, both experimental findings and theoretical analysis suggest that it is not as effective as our splatting-based approach. \textbf{Additional details are provided in the supplementary materials.}

The newly introduced velocity attributes of Gaussians can be optimized via backpropagation through the rasterization process. 
To utilize this property and optimize the velocity on Gaussians, we design specific losses for velocity field rendering to complement the conventional photometric loss.

\subsection{Losses on Rendered Velocity}
\textbf{Windowed Velocity Error ($\mathcal{L}_{\mathrm{win}}$).}
To refine the rendered velocity field at each iteration, we employ a temporal sliding window to sample multiple video frames during training. This approach ensures sufficient supervision and improves the model's consistency in predicting point cloud deformations across different timestamps. 
\textit{This sliding window operation is applied to all losses in this subsection.}
Specifically, given the rendered velocity field image $\hat{\mathbf{V}}_i$ at timestamp $i$, we perform $\tau+1$ rendering processes within the same iteration, which generates $\tau$ additional rendered velocity images, denoted as $\hat{\mathbf{V}}_i(\tau)=\{\mathbf{V}_{i+k}\}_{k=1}^{\tau}$, that span $k$ timestamps after $i$.
The windowed velocity error ($\mathcal{L}_{\mathrm{win}}$) is calculated between $\hat{\mathbf{V}}_i(\tau)$ and ground truth optical flows $\widetilde{\mathbf{V}}_i(\tau)$ at those timestamps. Formally: 
\begin{align} \label{eq:wve}
    \mathcal{L}_{\mathrm{win}} = \| \hat{\mathbf{V}}_i(\tau) - \widetilde{\mathbf{V}}_i(\tau) \|_1.
\end{align}
This loss ensures that the predicted velocity aligns with the actual scene dynamics across multiple consecutive frames.

\textbf{Flow Warping Error ($\mathcal{L}_{\mathrm{warp}}$).}
During training, the rendered velocity field registers the correspondence between RGB values of adjacent frames. To enhance this temporal correspondence, we propose the flow warping error ($\mathcal{L}_{\mathrm{warp}}$), which supervises frame-wise alignment by warping rendered RGB values in the next frame to the current frame using the rendered velocity field and minimizing its discrepancy from ground truth, as defined by:
\begin{align}
    \mathcal{L}_{\mathrm{warp}} = \| W_{{t+\delta t \rightarrow t}}(\hat{\mathbf{I}}_{t+\delta t}, \hat{\mathbf{V}}_{t+\delta t \rightarrow t}) - \tilde{\mathbf{I}}_t \|_1,
\end{align}
where $\hat{\mathbf{I}}$ and $\tilde{\mathbf{I}}$ represent rendered and ground truth images respectively, and $W_{t+\delta t \rightarrow t}$ denotes the warping operation from timestamp $t+\delta t$ to $t$. $\mathcal{L}_{\mathrm{warp}}$ provides another pathway of aligning the predicted velocity field with the actual motion dynamics observed in the scene, and helps to rectify any inaccuracies in the predicted motion, thereby improving the consistency of the deformation field and enhancing the overall fidelity of reconstruction.

\textbf{Dynamic Rendering Loss ($\mathcal{L}_{\mathrm{dyn}}$).}
To improve the rendering accuracy of dynamic objects in the scene, we perform dynamic foreground segmentation to isolate moving objects from the background and compute masked image error:
\begin{align}
    \mathcal{L}_{\mathrm{dyn}} = \| \hat{\mathbf{I} \odot \mathbf{M}_{\text{dyn}}}, \, \tilde{\mathbf{I}} \odot \mathbf{M}_{\text{dyn}} \|_1,
\end{align}
where $\mathbf{M}_{\text{dyn}}$ represents the dynamic foreground segmentation mask, and $\odot$ denotes element-wise multiplication. This loss emphasizes dynamic regions in the scene and focuses the optimization process on the motion and appearance of moving objects.
\begin{figure}[t]
  \centering
   \includegraphics[width=0.45\linewidth]{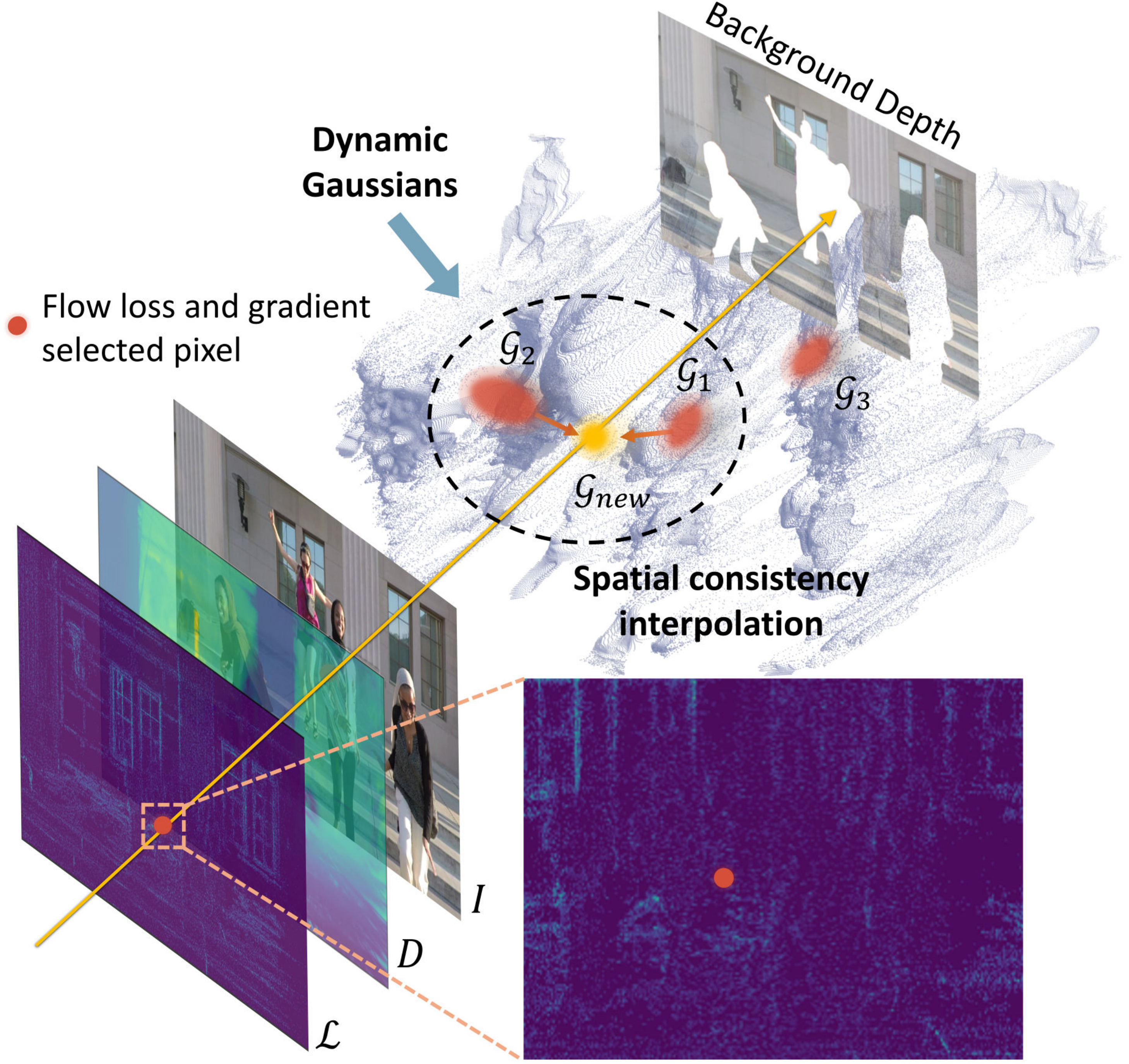}
   \vspace{-0.3cm}
   \caption{\textbf{Flow-assisted adaptive densification (FAD)} incorporates flow gradients and loss maps to identify pixel locations requiring densification. These pixels are lifted to the current frame's 3D space using rendered depth and then transformed back into the canonical space via deformation-induced displacement. The attributes of these new Gaussians are calculated through kNN-weighted interpolation.}
   \label{fig:fad}
   \vspace{-0.4cm}
\end{figure}

\subsection{Flow-assisted Adaptive Densification (FAD)} \label{subsec: fad}
Examining the experimental results of the baseline reveals its inability to handle challenging scenarios with significant object dynamics, where Gaussians are often absent in such regions. 
While the rendering and supervision of the velocity field effectively influence the positions and parameters of existing Gaussians, regions without Gaussians remain unoptimized. 
To address this limitation, we retain the 3DGS densification strategy while aiding it with the FAD strategy (\cref{fig:fad}). Unlike conventional densification that clones and splits existing Gaussians, FAD leverages flow-based losses and their gradients to directly add Gaussians to dynamic regions in the canonical space.
It aims to compensate for inadequate Gaussians in dynamic objects by aligning their positions with underlying motion dynamics, thereby achieving more accurate and temporally consistent representations. Specifically, it lifts prominent pixels in the velocity field, which are correlated to dynamic scene elements, into the 3D space and constructs 3D Gaussians of dynamic objects correspondingly, ultimately enhancing the model's representation and alignment with scene motion.

\textbf{Flow-assisted Pixel Selection.} We use the loss map $\mathcal{L}_{\mathrm{win}}$ and the gradient map $\nabla \mathcal{L}_{\mathrm{win}}$ from \cref{subsec:velocity fieid rendering} to identify pixel positions where additional Gaussians are needed.
We define a threshold $\epsilon$ to select pixel locations that exhibit sufficiently large losses and gradients: $\mathcal{P} = \{\mathbf{p}: \mathcal{L}_{\mathrm{win}}(\mathbf{p}) > \epsilon \text{ and } \nabla \mathcal{L}_{\mathrm{win}}(\mathbf{p}) > \epsilon\}$.
Since the loss and gradient maps may contain non-zero pixels in the background, we employ an off-the-shelf segmentation model SAM-v2~\cite{ravi2024sam2} to extract and exclude background pixels from $\mathcal{P}$.

\textbf{Lifting to 3D Space.} We lift the filtered pixels to 3D world coordinates $\mathcal{P}'$ using the rendered depth map $\mathbf{Z}$, camera intrinsics $\mathbf{K}$, and camera extrinsics [$\mathbf{R}$, $\mathbf{T}$], inverting the procedure described in Sec.~3.2. To lift a point $(x, y)$ from 2D space to 3D space, we use the intrinsic and extrinsic camera parameters, as well as the projection model. The general relationship between the 2D image coordinates and the 3D world coordinates can be expressed as:
\begin{align}
    \mathcal{P}' = (x_{3D},y_{3D},z_{3D})^T=\mathbf{K} [\mathbf{R}, \mathbf{T}] \cdot (x_{2D}, y_{2D}, 1)^T \cdot Z ,
\end{align}
where $(x_{2D}, y_{2D})$ are the 2D pixel coordinates in the image plane, $(x_{3D}, y_{3D}, z_{3D})$ are the corresponding 3D coordinates in the world space, $\mathbf{K}$ is the camera intrinsic matrix, which includes parameters such as focal length and principal point, $Z$ is the depth value corresponding to the point in the 3D world.

This equation assumes that we have the depth value $Z$ for the 3D point, which can be obtained through depth sensing or estimated through stereo vision techniques. Once the 3D coordinates are determined, we can perform operations such as motion tracking and dynamic scene reconstruction in 3D space.

\textbf{Farthest Point Sampling (FPS).} To keep the number of added points manageable, we employ the established FPS~\cite{eldar1997farthest,lang2020samplenet} to downsample the candidates with a ratio $r$. FPS works by iteratively selecting points that are farthest from those already selected in a point cloud. This process ensures that the selected subset of candidates, denoted as $\mathcal{P}''$, spreads evenly across the space.

\textbf{Spatial Consistency Interpolation.}
After downsampling, to ensure consistency of the new Gaussians' attributes in the canonical space, we use the K-nearest neighbors (kNN) algorithm to find the set of existing Gaussians surrounding each candidate in $\mathcal{P}''$.
To mitigate the spatial discontinuity of Gaussians parameters, we further obtain the radius $r_c$ on the pixel plane for each new Gaussian using differentiable rasterization and lift it to 3D to determine whether the neighboring points found by kNN lie within the candidate Gaussian’s spatial range. 
Points outside the radius indicate that the attributes between two Gaussians may not be continuous and thus cannot be used for attribute interpolation. 
To be more specific, we know that the FAD process operates on the frame at the current timestamp during training. 
The deformation network $D$ is achieved by transforming the point cloud $\mathbf{P}_0$ in the canonical space to obtain the current point cloud $\mathbf{P}_t$.
\begin{align}
    \mathbf{P}_t = \mathbf{P}_0 + D(\mathbf{P}_0, t).
\end{align}
Given that the pixels selected by $\mathcal{L}_{warp}$ and $\mathcal{L}_{win}$ are projected into the 3D space through depth to obtain new Gaussians $\mathcal{G}_f$, we use kNN to find the top-k nearest neighbors from the original Gaussians surrounding $\mathcal{G}_f$. The process can be described as follows:
\begin{align} \label{eq:knn}
    \mathcal{G}_c = \text{kNN}(\mathbf{P}_t, \mathcal{G}_f, k), \, \text{if} \, |\mathbf{P}_t - \mathcal{G}_f | < r_c,
\end{align}
where $r_c$ refers to the 3D range of each Gaussian in $\mathcal{G}_f$ and we finally obtain $\mathcal{G}_c$ as newly densified Gaussians.

Therefore, the new Gaussians $\mathcal{G}_c$ added at the current moment need to be mapped back to the canonical space $\mathcal{G}_c^{\text{cano}}$ by reversing the offset calculated through the deformation:
\begin{align}
    \mathcal{G}_c^{\text{cano}} = \mathcal{G}_c - D(\mathbf{P}_0, t).
\end{align}
\begin{table*}[ht]
\footnotesize
\caption{\textbf{Quantitative evaluation on the Nvidia dataset}~\cite{yoon2020dynamicnvs}. The resolution and video length of each scene vary. ``Face.'' and ``Pg.'' refer to dynamicFace and Playground. {[PSNR (DPSNR)$\uparrow$, SSIM$\uparrow$, LPIPS$\downarrow$]} are reported. Same applies to \cref{tab:neu3d-all-scene-1}.
}
\centering
\adjustbox{width=1.0\linewidth}{
\begin{tabular}{l|ccccc}
\toprule
Scene & \textbf{4DGS} & \textbf{4D-GS} & \textbf{SC-GS} & \textbf{MotionGS} & \textbf{Ours} \\ 
\midrule
Balloon1 & 20.51 (21.24) / 0.619 / 0.317 & 23.06 (21.78) / 0.740 / 0.210 & 18.17 (16.94) / 0.666 / 0.440 & 20.72 (19.97) / 0.611 / 0.472 & \textbf{24.50 (24.54)} / 0.757 / 0.290 \\
Skating & 26.80 (16.35) / 0.885 / 0.203 & 28.06 (18.32) / 0.872 / 0.131 & 18.09 (7.86) / 0.529 / 0.535 & 23.21 (12.94) / 0.834 / 0.370 & \textbf{29.91 (19.56)} / 0.916 / 0.180 \\
Face. & 19.47 (24.58) / 0.783 / 0.185 & 18.61 (22.40) / 0.801 / 0.153 & 12.38 (9.25) / 0.332 / 0.546 & 15.15 (20.67) / 0.560 / 0.446 & \textbf{21.82 (27.20)} / 0.854 / 0.042 \\
Jumping & 24.33 (17.75) / 0.847 / 0.246 & 23.80 (18.25) / 0.803 / 0.199 & 13.37 (7.96) / 0.368 / 0.639 & 20.53 (14.31) / 0.771 / 0.403 & \textbf{27.89 (21.07)} / 0.887 / 0.216 \\
Truck & 25.18 (22.19) / 0.805 / 0.279 & 24.11 (20.67) / 0.791 / 0.260 & 16.87 (12.95) / 0.486 / 0.612 & 20.47 (18.68) / 0.698 / 0.498 & \textbf{26.22 (23.96)} / 0.835 / 0.255 \\
Pg. & 19.98 (16.65) / 0.602 / 0.319 & 21.66 (16.92) / 0.757 / 0.179 & 12.27 (13.12) / 0.197 / 0.613 & 18.07 (15.62) / 0.448 / 0.533 & \textbf{22.18 (18.15)} / 0.695 / 0.281 \\
Umbrella & 22.85 (21.24) / 0.527 / 0.416 & 24.01 (21.84) / 0.574 / 0.346 & 6.76 (4.65) / 0.000 / 0.719 & 20.80 (19.19) / 0.471 / 0.625 & \textbf{24.11 (22.32)} / 0.599 / 0.366 \\
\midrule 
\textbf{Avg.} & 22.73 (20.00) / 0.724 / 0.280 & 23.33 (20.03) / 0.763 / 0.211 & 13.99 (10.39) / 0.368 / 0.586 & 19.85 (17.34) / 0.628 / 0.478 & \textbf{25.23 (22.40)} / 0.792 / 0.232 \\
\bottomrule
\end{tabular}
\vspace{-0.3cm}
}
\label{tab:nvidia-all-scene-1}
\end{table*}
\begin{table*}[ht]
\footnotesize
\caption{\textbf{Quantitative evaluation on the Neu3D dataset}~\cite{li2022neu3d} with a resolution of 1,352$\times$1,014.}
\centering
\adjustbox{width=1.0\linewidth}{
\begin{tabular}{l|ccccc}
\toprule
Scene & \textbf{4DGS} & \textbf{4D-GS} & \textbf{SC-GS} & \textbf{MotionGS} & \textbf{Ours} \\ 
\midrule
martini & 22.47 (20.97) / 0.810 / 0.240 & 23.85 (21.22) / 0.861 / 0.219 & 13.33 (9.24) / 0.306 / 0.615 & 21.05 (20.17) / 0.805 / 0.405 & \textbf{24.19 (23.69)} / 0.856 / 0.206 \\
spinach & 24.92 (22.88) / 0.831 / 0.222 & 26.81 (26.76) / 0.833 / 0.261 & 17.11 (15.89) / 0.432 / 0.531 & 25.94 (23.20) / 0.864 / 0.333 & \textbf{27.06 (25.40)} / 0.881 / 0.186 \\
beef & 26.61 (25.08) / 0.881 / 0.192 & 28.48 (28.79) / 0.916 / 0.163 & 16.42 (15.74) / 0.414 / 0.552 & 21.55 (20.37) / 0.832 / 0.367 & \textbf{29.46 (28.58)} / 0.924 / 0.166 \\
salmon & 22.95 (21.45) / 0.844 / 0.209 & 24.79 (23.24) / 0.896 / 0.171 & 8.23 (6.79) / 0.186 / 0.650 & 22.49 (21.62) / 0.813 / 0.388 & \textbf{25.67 (24.79)} / 0.883 / 0.180 \\
flame & 25.70 (23.49) / 0.851 / 0.218 & \textbf{29.35} (26.37) / 0.951 / 0.143 & 6.71 (5.98) / 0.054 / 0.654 & 25.92 (23.04) / 0.878 / 0.323 & 28.86 (\textbf{26.59}) / 0.927 / 0.157 \\
sear & 26.46 (26.12) / 0.868 / 0.199 & 26.94 (29.36) / 0.914 / 0.163 & 6.72 (6.39) / 0.052 / 0.660 & 24.96 (26.13) / 0.865 / 0.347 & \textbf{28.54 (29.49)} / 0.922 / 0.156 \\
\midrule 
\textbf{Avg.} & 24.85 (23.33) / 0.847 / 0.213 & 26.70 (25.96) / 0.895 / 0.187 & 11.42 (10.01) / 0.241 / 0.610 & 23.65 (22.42) / 0.982 / 0.361 & \textbf{27.30 (26.47)} / 0.899 / 0.175 \\
\bottomrule
\end{tabular}
}
\label{tab:neu3d-all-scene-1}
\vspace{-0.2cm}
\end{table*}

\section{Temporal Velocity Refinement (TVR) with Extended Kalman Filtering (EKF)}
\label{subsec:temporal_refinement}

While deformation network-based methods can establish temporal correlations, they do not inherently ensure smooth and physically consistent motion. In light of this, we here propose a \emph{temporal velocity refinement (TVR)} process following the “refinement as a post-processing step” philosophy, whose impact on motion prediction will be explored in the subsequent experiment section.

\textbf{Extended Kalman Filtering.} As an optimal recursive estimation technique in control theory, Kalman filter (KF)~\cite{wan2000unscented,zhan2024kfd} estimates the internal states $\mathbf{x}$ of a linear dynamic system in the presence of noisy observations $\mathbf{z}$.
The system is described using the state-space equations below:
\begin{align}
    \begin{split}
        \mathbf{x}_{k} &= \mathbf{A} \mathbf{x}_{k-1} + \mathbf{B} \mathbf{u}_{k} + \mathbf{w}_{k},\\ \mathbf{z}_{k} &= \mathbf{H} \mathbf{x}_{k} + \mathbf{v}_{k},
    \end{split}
\end{align}
where $\mathbf{A}$ and $\mathbf{H}$ are the state transition matrix and the observation matrix, and $\mathbf{w}_{k} \sim \mathcal{N}(0, \mathbf{Q})$ and $\mathbf{v}_{k} \sim \mathcal{N}(0, \mathbf{R})$ are noise terms. For the task of temporal velocity refinement, the input term $\mathbf{B}\mathbf{u}_k$ is not needed and is omitted hereafter.
To generalize this formulation to nonlinear systems, extended Kalman filter~\cite{terejanu2008extended,kang2016adaptive} replaces $\mathbf{A}$ and $\mathbf{H}$ with nonlinear functions $\mathbf{f}$ and $\mathbf{h}$ whose local linearity at $\mathbf{x}$ is characterized by Jacobian matrices $\mathbf{J}_\mathbf{f}(\mathbf{x})$ and $\mathbf{J}_\mathbf{h}(\mathbf{x})$ respectively. 
Then, we iterate between the \emph{forecast step} and the \emph{assimilation step} to progressively refine $\mathbf{x}_k$ given previous estimation $\mathbf{x}_{k-1}$ and current observation $\mathbf{z}_k$ for each time step $k$. More details on EKF are provided in the \textbf{supplementary materials}.

Intuitively, EKF imposes a stronger correction of $\mathbf{x}_k^f = \mathbf{f}(\mathbf{x}_{k-1})$ when the accumulated covariance $\mathbf{P}_k$ grows larger, thereby smoothing the state sequence and reducing noisy or zigzagging patterns. Therefore, it stands as a promising tool for the refinement of Gaussian velocity.
Applying EKF to the \model pipeline, we define
$\mathbf{f}$ and $\mathbf{h}$ as:
\begin{align}
    \begin{split}
        \mathbf{f}(\mathbf{x}_{t-1}) &= D(D^{-1}(\mathbf{x}_{t-1}, t-1), t),\\
        \mathbf{h}(\mathbf{x}) &= \mathrm{VRaster}(\mathbf{x}),
    \end{split}
\end{align}
where $D$ is the deformation network and $\mathrm{VRaster}(\cdot)$ is the velocity rasterization process described \cref{subsec: prelim}.
Since both $D$ and $\mathrm{VRaster}(\cdot)$ are intractable, we have to assume their local linearity and approximate the partial gradients as:
\begin{align}
    \begin{split}
    \mathbf{J}_{\mathbf{f}} (\mathbf{x}_{t-1}) &= \frac{\partial \mathbf{f}(\mathbf{x}_{t-1})}{\partial \mathbf{x}} \approx  \frac{ \mathbf{f}(\mathbf{x}_{t-1}+\delta \mathbf{x}) - \mathbf{x}_{t}}{\delta \mathbf{x}},\\ 
        \mathbf{J}_{\mathbf{h}} (\mathbf{x}_{t}) &= \frac{\partial \mathbf{h}(\mathbf{x}_t)}{\partial \mathbf{x}} \approx \frac{\mathbf{h}(\mathbf{x}_{t}+\delta \mathbf{x}) - \mathbf{h}(\mathbf{x}_{t})}{\delta \mathbf{x}}.
    \end{split}
\end{align}
Furthermore, since $D$ is non-invertible and directly obtaining $\mathbf{f}(\mathbf{x}_{t-1} + \delta\mathbf{x})$ is impossible, we take a detour by considering the Jacobian of $D$ ($\mathbf{J}_D$): $\mathbf{J}_\mathbf{f}(\mathbf{x}_{t-1}) = \mathbf{J}_D(\mathbf{x}_t)\cdot\mathbf{J}_D(\mathbf{x}_{t-1})^{-1}$ from the inverse function theorem. A more detailed calculation is provided in the \textbf{supplementary materials}.

\textbf{Surface Gaussian filtering.} 
In an image rendered from 3D Gaussians, multiple Gaussians could project to the same pixel, yet only those on the surface of objects contribute to the optical flow value of that pixel. Therefore, we filter out each Gaussian whose projected depth is greater than the rendered depth value of its corresponding pixel at each time step, so that occluded Gaussians are not mistakenly updated. Note that we still run the forecast step for those excluded Gaussians, while skipping the assimilation step by directly using $\mathbf{x}_k^f$ and $\mathbf{P}_k^f$ as $\mathbf{x}_k$ and $\mathbf{P}_k$.

\textbf{Accurate flow map localization.} 
Another challenge of TVR is to select the correct optical flow observation for each Gaussian. While using the projected coordinates at each time step sounds plausible, this approach links the reliability of $\mathbf{z}_k$ to the accuracy of $\mathbf{x}_{k-1}$. That is, for an erroneous $\mathbf{x}_{k-1}$, the selected observation will also deviate from the ground truth flow value, which further causes the assimilated $\mathbf{x}_k$ to have an even larger error and finally leads to catastrophic error accumulation. 
To resolve this problem, we assume that Gaussian locations are reliable only at the first frame, and for the $k^{\text{th}}$ time step, $\mathbf{z}_k$ is calculated as: $\mathbf{z}_k = \mathbf{F}(\mathbf{p}_0 + \sum_{i=0}^{k-1}\mathbf{z}_i)$, where $\mathbf{F}(\mathbf{p})$ denotes the optical flow value at coordinates $\mathbf{p}$, and $\mathbf{p}_0$ is the projected 2D location of $\mathbf{x}_0$. That is, the coordinates used to index the flow map from the second frame on are obtained via adding up the flow values from all previous time steps.

\begin{figure*}[ht]
    \centering
    \includegraphics[width=1.0\linewidth]{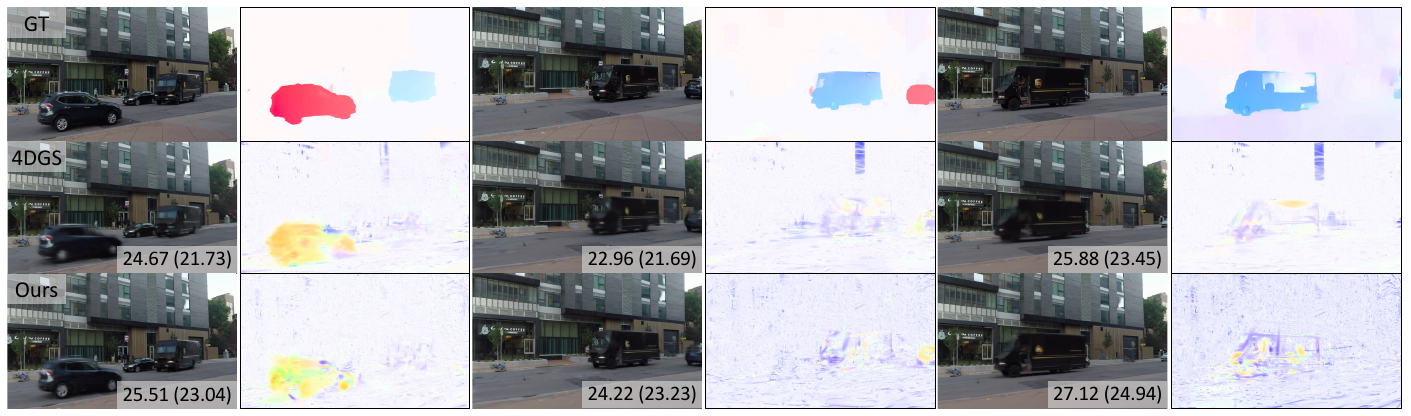}
    \vspace{-0.5cm}
    \caption{\textbf{Novel view synthesis evaluation on the Nvidia ``Truck'' scene}. We compare our work with the retrained baseline (4DGS) over a few timesteps. As each timestep, we show the rendered novel view image with PSNR (DPSNR) in dB, and the estimated velocity field (color gamma-tuned for visualization).}
    \label{fig:sota nvidia-long}
    \vspace{-0.6cm}
\end{figure*}

\begin{figure}[ht]
  \centering
   \includegraphics[width=0.5\linewidth]{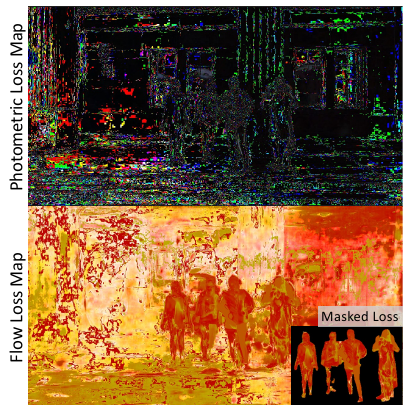}
   \vspace{-0.3cm}
   \caption{\textbf{Comparison of photometric and flow-based losses.} We visualize the pixel-wise photometric loss and our proposed flow-based loss for a given frame. We observe that $\mathcal{L}_{\mathrm{warp}}$ and $\mathcal{L}_{\mathrm{win}}$ are primarily concentrated on foreground pixels.}
   \label{fig:compare gradient}
   \vspace{-0.5cm}
\end{figure}

\begin{figure}[h]
  \centering
   \includegraphics[width=0.5\linewidth]{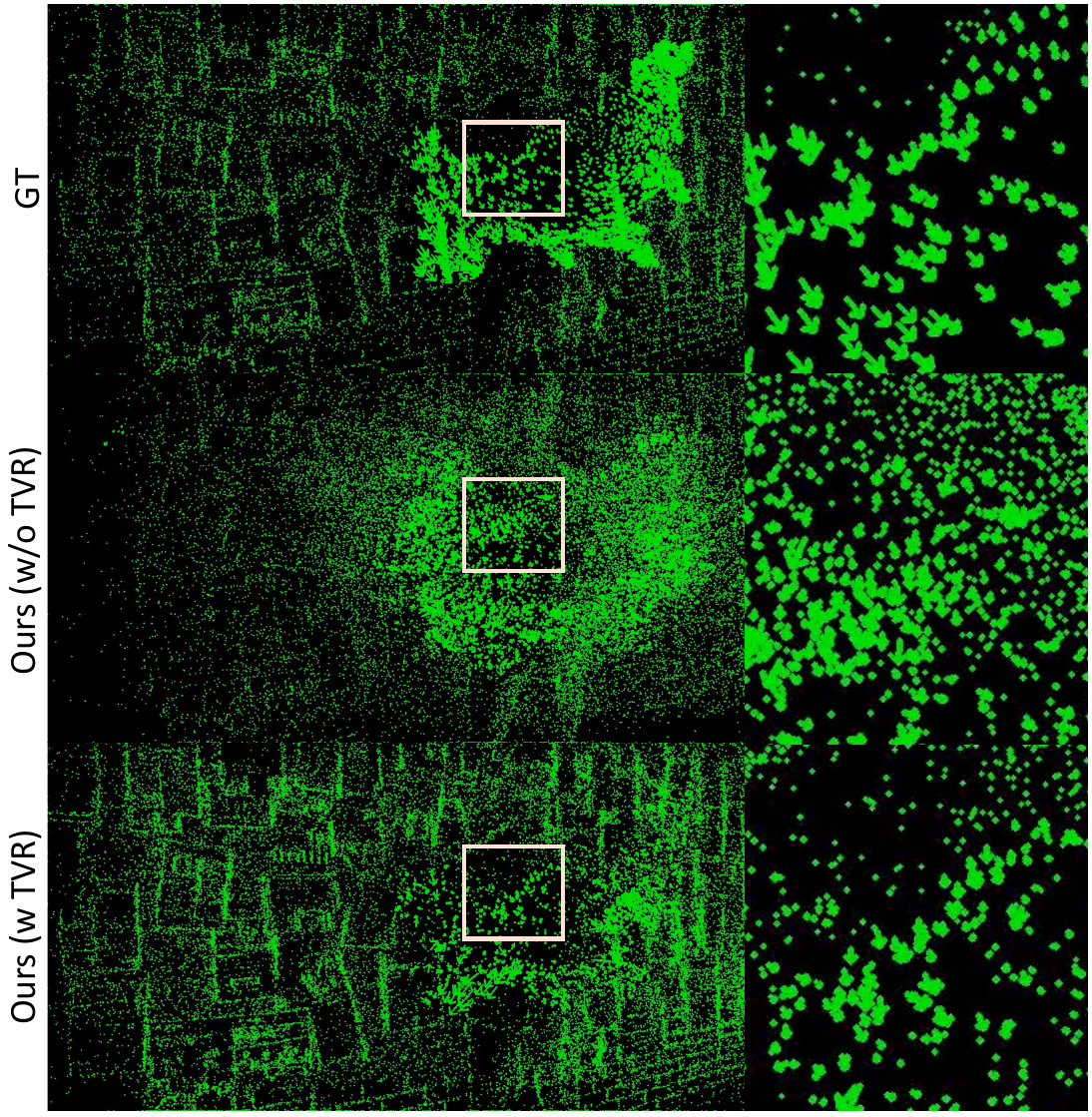}
   \vspace{-0.3cm}
   \caption{Visualization of the effectiveness of Temporal Velocity Refinement (TVR) trajectory correction. The figure depicts the 3D Gaussian motion rendered as a 2D velocity field, with arrows indicating the velocity vectors between consecutive frames.}
   \label{fig:ekf viz}
   \vspace{-0.5cm}
\end{figure}

\begin{figure*}[p]
    \centering
    \includegraphics[width=1.0\linewidth]{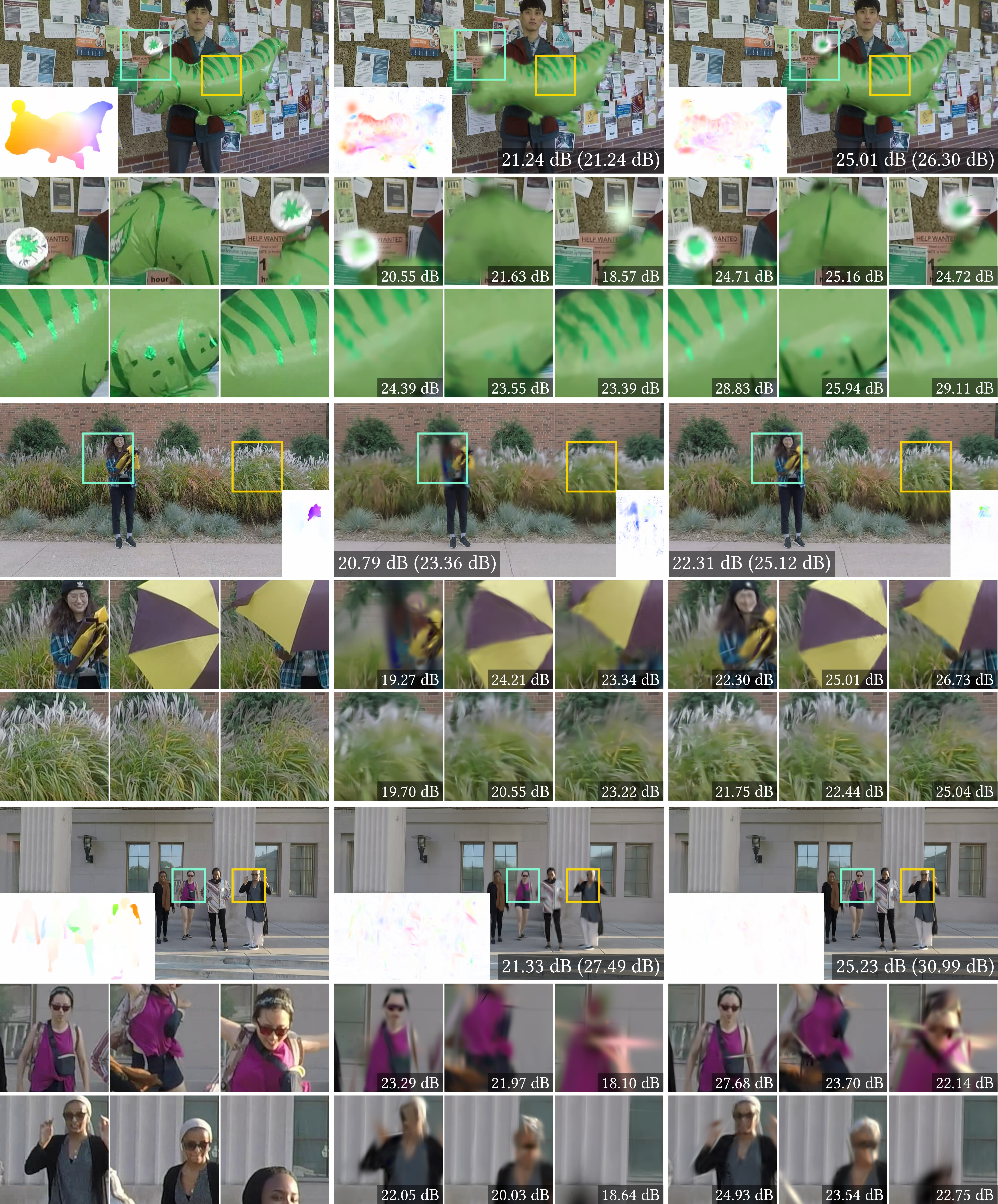}
    \caption{\textbf{Novel view synthesis evaluation on the Nvidia dataset}. We present three scenes: ``Balloon1'', ``Umbrella'', and ``Jumping'' and compare our work with retrained 4DGS. The 3 primary columns show results of the ground truth, baseline, and our method. For each scene, ground truth optical flows (first column) and rendered velocity fields (other columns) are presented in false color. The second and third rows within each scene show insets of 2 subareas, from left to right yielding 3 different timestamps. The PSNR showing at corners is calculated using the entire image.}
    \label{app fig:sota nvidia}
\end{figure*}

\begin{figure*}[h]
    \centering
    \includegraphics[width=1.0\linewidth]{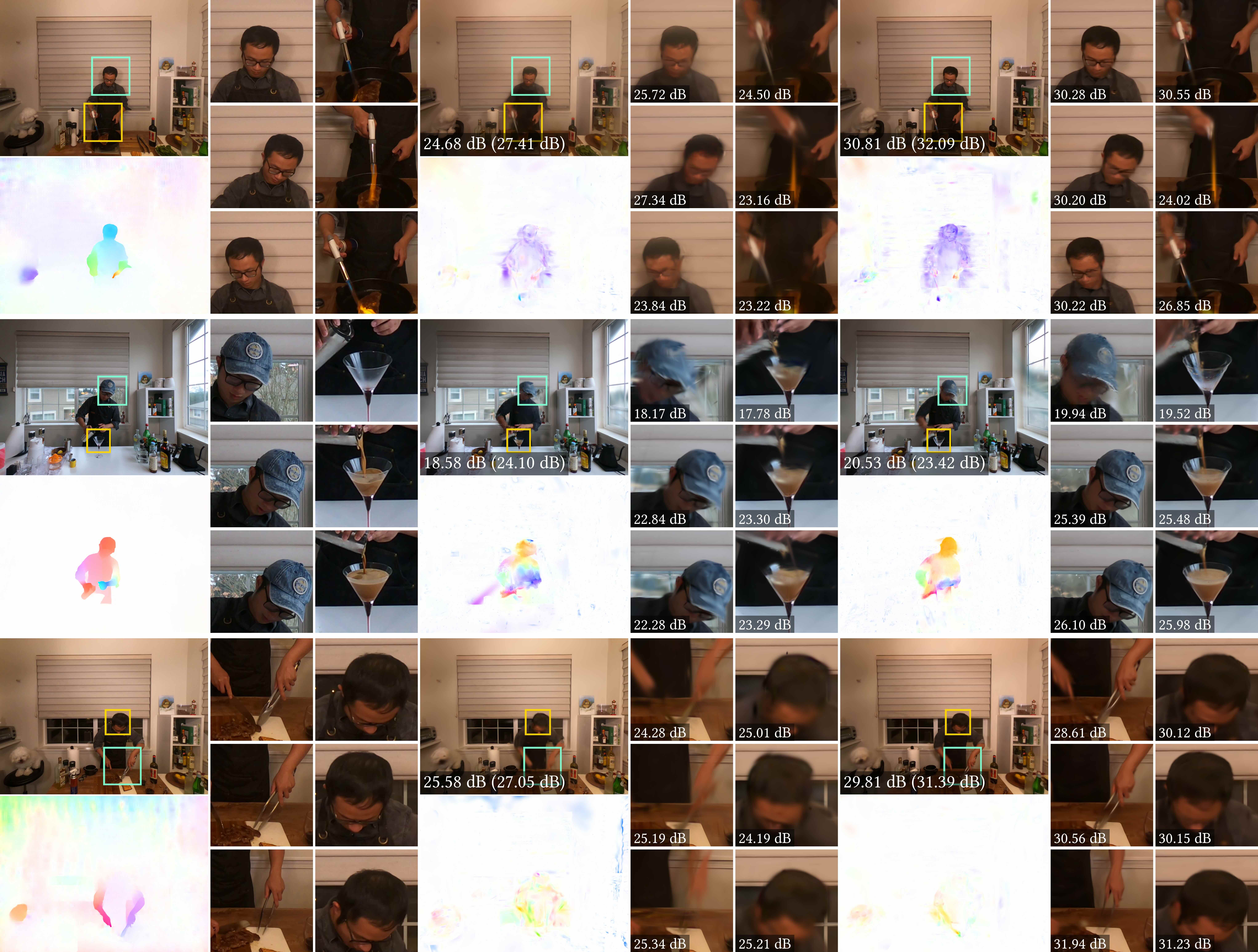}
    \caption{\textbf{Novel view synthesis evaluation on the Neu3d dataset}. We visualize the ``flame'', ``martini'', and ``beef'' scenes and compare our approach with the retrained 4DGS. We rearranged the inset placement from figure \ref{app fig:sota nvidia} to suit the aspect ratio of this dataset, that the 3 main columns from left to
    right are still the results of the ground truth, 4DGS, and our method, and the insets from top to bottoms within each group represents 3 different time stamps. The colored image on the bottom-left of each group is optical flow for the ground truth, and velocity field rendering for the others accordingly.}
    \label{app fig:sota neu3d}
    \vspace{-0.3cm}
\end{figure*}

\section{Experiment}
\label{sec:experiments}

\subsection{Initialization and Implementation Details}
Similar to the initialization strategy in 3DGS~\cite{kerbl3Dgaussians}, reconstruction quality improves when the initial scene point cloud is derived from SfM~\cite{schonberger2016sfm}. Following 4DGS~\cite{wu20244d}, the first frame of each video sequence is used as the initialization frame, and COLMAP~\cite{schonberger2016colmap} is applied to generate Gaussian centers. Ground truth optical flows are obtained using RAFT~\cite{teed2020raft}, and dynamic foreground segmentation masks are manually selected with SAM-2~\cite{ravi2024sam2}.

We conducted quantitative evaluations on the Nvidia-long~\cite{yoon2020dynamicnvs} dataset, a benchmark for dynamic scenes, pre-processed and made publicly available by prior works~\cite{li2023dynibar}. This dataset includes 7 scene videos, each containing 90–210 frames with 12 camera views per frame. The Neu3D dataset~\cite{li2022neu3d}, captured with 15–20 static cameras over extended periods with complex motions, features 6 selected scene sequences of 300 frames each. These datasets comprise high-resolution videos of dynamic scenes, offering challenges such as occlusions, rapid motion, and large-scale changes.

Our \model pipeline builds upon 4DGS~\cite{Wu_2024_CVPR}, and compares with 4DGS, 4D-GS~\cite{yang2023gs4d}, SC-GS~\cite{huang2024scgs}, and MotionGS~\cite{zhu2024motiongs}. All training is performed on an NVIDIA RTX 4090 GPU. More experimental details can be found in the \textbf{supplementary material.} Notably, we set the FPS point cloud downsampling ratio in FAD (Sec~\ref{subsec: fad}) to 0.01 for Nvidia-long and 0.001 for Neu3D.

\subsection{Results}
The quantitative results on the Nvidia-long~\cite{yoon2020dynamicnvs} and Neu3D~\cite{li2022neu3d} datasets are summarized in \cref{tab:nvidia-all-scene-1} and \cref{tab:neu3d-all-scene-1}, respectively. A qualitative comparison between \model and the 4DGS baseline can be found in \cref{app fig:sota nvidia} and \cref{app fig:sota neu3d}. In addition to standard metrics such as PSNR, SSIM, and LPIPS, we report dynamic PSNR (DPSNR) to specifically evaluate the rendering quality of dynamic foreground pixels. 
D-PSNR computes PSNR only in dynamic regions, identified by SAM-2, to assess reconstruction quality in motion areas.
Our experimental results demonstrate that \model significantly outperforms all three baselines in novel view synthesis. The average PSNR metrics across all scenes in the Neu3D and Nvidia-long datasets are improved by 2.45 dB and 2.5 dB, respectively, compared to the best-performing 4DGS method. For dynamic scene rendering, our method achieves average improvements of 3.14 dB and 2.4 dB in DPSNR over 4DGS.
These metrics validate that our method effectively optimizes both static and dynamic components of the scene, and the gain in DPSNR is more pronounced compared to the overall PSNR, highlighting the effectiveness of \model on capturing and reconstructing dynamic scene contents.

\begin{table}[ht]
\footnotesize
    \centering
    \caption{\textbf{Effectiveness assessment of Velocity Rendering (V.R.), FAD, $\mathcal{L}_{\mathrm{warp}}$, and $\mathcal{L}_{\mathrm{win}}$} on dynamic scene reconstruction. ``N'' refers to the number of Gaussians in the final point cloud.}
    \vspace{-2pt}
    \adjustbox{width=0.5\linewidth}{\begin{tabular}{l|cccc|c|c}
        \toprule
         & V.R. & FAD & $\mathcal{L}_{\mathrm{warp}}$ & $\mathcal{L}_{\mathrm{win}}$ & PSNR / SSIM / LPIPS & N \\
        \cmidrule(lr){1-7}
        \multirow{4}{*}{\rotatebox{90}{4DGS}} & \xmark & \xmark & \xmark & \xmark &  20.51 / 0.619 / 0.317 & 214k \\
        & \cmark & \cmark & \cmark & \xmark & 22.69 / 0.703 / 0.312 & 89k \\
        & \cmark & \cmark & \xmark & \cmark & 22.47 / 0.718 / 0.305 & 75k \\
        & \cmark & \cmark & \cmark & \cmark & \cellcolor{best} 24.50 / 0.757 / 0.290 & 141k \\
        \bottomrule
    \end{tabular}}
    \label{tab:abl-1}
    \vspace{-0.4cm}
\end{table}
\begin{table}[h]
\centering
\footnotesize
\caption{\textbf{Effectiveness assessment of sliding window size for $\mathcal{L}_{\mathrm{win}}$} on final rendering quality. Metric reported is PSNR (dB)$\uparrow$.}
\label{tab:abl num adj}
\begin{tabular}{l|cccc}
\toprule
Number of adjacent frames & \bf{2} (4DGS) & \bf{4} & \bf{6} & \bf{8} \\
\midrule
Jumping & 25.33 & 25.63 & 26.45 & \cellcolor{best} 27.89 \\
Coffee-martini & 22.47 & 22.68 & 23.23 & \cellcolor{best} 24.19 \\
\bottomrule
\end{tabular}
\vspace{-0.25cm}
\end{table}

We further provide a close-up inspection of velocity fields for synthesized novel views on the ``Truck'' scene to demonstrate the effectiveness of our method in \cref{fig:sota nvidia-long}. 
In this scene, the primary reconstruction challenges lie in the significant scale variations between the truck and the car, motion blur of the car, and fine-grained details such as text on the truck. As shown in the rendered Gaussian velocity field, our method produces gradually enlarging Gaussians on the truck’s front as it approaches, aligning well with the physical motion. In contrast, 4DGS introduces inconsistent Gaussian replacements that violate geometric continuity. Additionally, the RGB rendering results further confirm that Gaussians adhering to plausible motion dynamics yield more realistic visual quality.

\subsection{Ablation Study} 
\label{sec:ablation}
For ablation studies, we chose one representative scene from each dataset based on the level of difficulty: the relative easy ``martini'' from Neu3D and the hard ``jumping'' from Nvidia-long. 
We validate the effectiveness of velocity rendering, $\mathcal{L}_{\mathrm{warp}}$, and $\mathcal{L}_{\mathrm{win}}$ in reconstructing dynamic scenes and report the results in \cref{tab:abl-1}.

\textbf{Ablation on Sliding Window Size.}
We study the impact of the sliding window size in $\mathcal{L}_{\mathrm{win}}$ on the reconstruction performance, and present the results in \cref{tab:abl num adj}.
We observe that a larger window size results in stronger temporal consistency and thus incurs more accurate velocity estimation of moving objects. Such benefits are attributed to the temporal window overlapping between adjacent iterations, bringing stability to the training process. 

\textbf{Loss Map Visualization.}
We visualize the loss maps of the $\mathcal{L}_1$ photometric loss and flow-based losses for a rendered frame in~\cref{fig:compare gradient}. We note that $\mathcal{L}_1$ does not emphasize dynamic regions and therefore misleads conventional densification algorithms to prioritize static backgrounds, potentially resulting in inaccurate reconstructions of dynamic elements. 
On the other hand, our $\mathcal{L}_{\mathrm{win}}$ and $\mathcal{L}_{\mathrm{warp}}$ generally yield larger loss values for dynamic pixels, making them more suitable for guiding the densification steps in dynamic scenes. 
By additionally applying a dynamic mask to filter out irrelevant losses from the background, we further ensure that all Gaussian candidates in FAD correspond to foreground objects.

\textbf{Evaluation of TVR.} By applying the temporal velocity refinement (TVR) with EKF to the ``Balloon1'' scene from Nvidia-long, we observe that the PSNR on novel views decreases from 24.50 to 24.32 (-0.18 change) and the DPSNR decreases from 24.54 to 24.17 (-0.37 change). While TVR induces a slight deterioration in render quality, we yet observe that the accuracy of Gaussians' offsets (motion), and subsequently the modeling of scene dynamics, has notably improved, as shown in \cref{fig:ekf viz}. We argue that the drop in visual quality is due to TVR not being able to optimize Gaussian attributes other than the center coordinates.

\section{Conclusion and Analysis}
\label{sec:conclusion}

In this work, we present an enhanced flow-based velocity field modeling technique to improve Gaussian deformation for video reconstruction. 
By expanding the capabilities of the 3DGS rasterizer to accommodate velocity field rendering, we introduce a velocity attribute to each Gaussian and incorporate multiple constraints --- $\mathcal{L}_{\mathrm{win}}$, $\mathcal{L}_{\mathrm{warp}}$. and $\mathcal{L}_{\mathrm{dyn}}$ --- to ensure temporal consistency and smoothness in both RGB rendering and trajectory predictions. 
Additionally, we develop FAD, a method that actively adds Gaussians in challenging dynamic regions.
Extensive experiments conducted on challenging datasets illustrate that our model demonstrates improved alignment with real motion and reduced motion blur, leading to clearer textures for smoother motions. 
We believe our approach holds significant potential for achieving more accurate and temporally consistent dynamic scene reconstruction in applications such as VR/AR, gaming, and cinematic content creation.

\textbf{Limitations and future work.} 
Our exploration of temporal velocity refinement using EKF is still in its infant stages, yet it has shown promise in refining Gaussian center trajectories alongside other attributes, and is worth further research and experimentation.
The current model encounters challenges in accurately predicting motion trajectories under complex and rapid motion, particularly in scenarios  involving abrupt object changes (e.g., abrupt appearance or size variations), for example, the ``Jumping'' scene in Nvidia-long. As shown in \cref{app fig:sota nvidia}, the visual quality and PSNR of reconstruction decrease for later timestamps, when persons in the scene move closer and existing Gaussians are no longer sufficient to fully represent their fine details.
Future improvements in dynamic Gaussian control and adaptive optimization strategies could possibly address these issues. 
Additionally, maintaining Gaussian attribute consistency across views during significant camera perspective changes (e.g., 360-degree rotations~\cite{Joo_2017_pano}) remains a challenge.
Lastly, the sliding window strategy for loss computation leads to increased training time.
Addressing these aspects presents opportunities for future improvements in velocity-based modeling.


\bibliographystyle{unsrt}  
\bibliography{references}

\end{document}